\pdfoutput=1

\documentclass[11pt]{article}

\usepackage[]{EMNLP2023}

\usepackage{times}
\usepackage{latexsym}
\usepackage{multirow}

\usepackage[T1]{fontenc}

\usepackage[utf8]{inputenc}

\usepackage{microtype}

\usepackage{inconsolata}

\usepackage{pgf}
\usepackage{tikz}
\usepackage{hhline}
\usepackage{collcell}
\usepackage{tikz}
\def\checkmark{\tikz\fill[scale=0.4](0,.35) -- (.25,0) -- (1,.7) -- (.25,.15) -- cycle;} 

\title{Pre-training LLMs using human-like development data corpus}

\makeatletter
\def\thanks#1{\protected@xdef\@thanks{\@thanks
        \protect\footnotetext{#1}}}
\makeatother

\newcommand{\gtlogo}{\raisebox{3.4pt}{\includegraphics[scale=0.04]{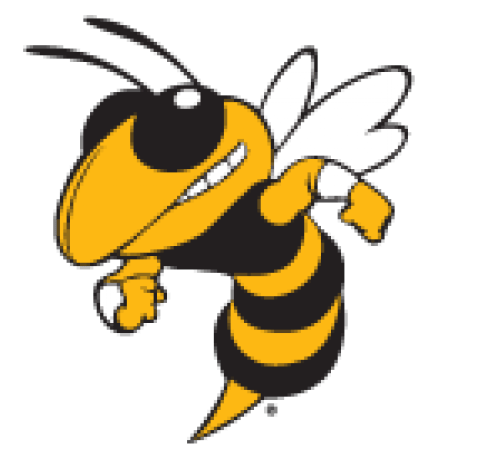}}}

\author{
Khushi Bhardwaj, Raj Sanjay Shah, Sashank Varma
 \\ 
 Georgia Institute of Technology \gtlogo \\
 \textcolor{darkblue}{\{\href{mailto:khushi.bhardwaj@gatech.edu}{khushi.bhardwaj}, 
 {\href{mailto:rajsanjayshah@gatech.edu}{rajsanjayshah},
\href{mailto:varma@gatech.edu}{varma}\}@gatech.edu}
}}

\begin{document}
\maketitle
\begin{abstract}
Pre-trained Large Language Models (LLMs) have shown success in a diverse set of language inference and understanding tasks. The pre-training stage of LLMs looks at a large corpus of raw textual data. The BabyLM shared task compares LLM pre-training to human language acquisition, where the number of tokens seen by 13-year-old kids is magnitudes smaller than the number of tokens seen by LLMs. In this work, we pre-train and evaluate LLMs on their ability to learn contextual word representations using roughly the same number of tokens as seen by children. We provide a strong set of baselines; with different architectures, evaluation of changes in performance across epochs, and reported pre-training metrics for the strict small and strict tracks of the task. We also try to loosely replicate the RoBERTa baseline given by the task organizers to observe the training robustness to hyperparameter selection and replicability. We provide the submission details to the strict and strict-small tracks in this report.
\end{abstract}

\section{Introduction}
Transformer-based LLMs \cite{transformers} show state-of-the-art performance on a variety of language processing tasks. In the last few years, pre-training methods for LLMs have evolved rapidly to meet task-driven demands. This evolution has focused on model expansion \cite{gpt3}, more pre-training data \cite{chinchilla}, use of higher quality data \cite{t5}, model alignment \cite{trx}, quicker run-time inference \cite{sanh2020distilbert}, quicker pre-training \cite{electra}, faster fine-tuning \cite{sanh2020distilbert}, domain adaptation \cite{clinicalbert, caselli2021hatebert, scibert, EMNLP_raj}, and the addition of multi-modal capabilities \cite{openai2023gpt4, gatti-etal-2022-vistot}. The task-driven nature of this development optimizes performance at scale but fails to account for human-like learning.

Humans typically encounter fewer than 100 million tokens through language exposure by the time they are 13 years old \cite{warstadt-et-al-2023-babylm}. LLMs, on the other hand, parse tens of billions to trillions of tokens in their pre-training stage, typically from sources like Wikipedia \cite{wiki}, and Open Book Corpus \cite{bookcorpus}, which consist of different tokens than the ones seen by children. In this paper, we evaluate the capabilities of popular architectures on various tasks when trained on a number of tokens comparable to that seen by 13-year-old children. Such scaled-down pre-training has several potential benefits: 
\begin{itemize}
    \item A better sandbox for the development of new LLM training techniques inspired by the cognitive science literature \cite{yiu2023imitation}.
    \item Robust evaluation of models on human behavioral signatures \cite{shah2023human}.
    \item Building plausible human cognition models using LLMs aligned to actual human actions \cite{park2022social}.
\end{itemize}
\begin{table}[!ht]
\resizebox{0.48\textwidth}{!}{%
    \centering
    \begin{tabular}{lp{1cm}p{3.5cm}c}
    \hline
        Track & Data size & Datasets & Our work 
        \\ \hline
        Strict-small & 10M words & Child-directed speech, transcribed speech from multiple sources, children’s books, and Wikipedia, etc. & \checkmark \\ 
        Strict & 100M words & ~ & \checkmark \\ \hline
        Loose & 100M words & Strict track data + unlimited non-linguistic data & $\times$ \\ \hline
    \end{tabular}
        }
    \caption{Task Summary }
    \label{tab:tasks}

\end{table}
\subsection{Task Descriptions}
The shared task has three tracks: Strict, Strict-small, and Loose. The details of each track are summarized in Table \ref{tab:tasks}. The Strict and Strict-small tracks use pre-released datasets containing Child-directed speech, transcribed speech from multiple sources, children’s books, and Wikipedia. These tracks are meant to encourage explorations of architectural variation and self-supervised approaches.

\subsection{Key Contributions}
Given the benefits of using scaled-down human-like pre-training data,  our work focuses on the following aspects of the shared task:
\begin{enumerate}
    \item Replication details: Can we replicate the results of the baselines given by the task organizers?
    \item Can we understand the impact of more training epochs on the same architecture?
    \item Providing each training checkpoint for the different model architectures to facilitate future modeling of development. All checkpoints can be found \textbf{\href{https://huggingface.co/Raj-Sanjay-Shah}{here.}}
\end{enumerate}
We provide details of training and evaluation for the strict and strict-small tracks of this task.
\section{Related Work}
\subsection{Cognitive science driven LLM architecture development}


With the efforts put into LM pre-training, learning frameworks informed by cognitive science have received increasing attention. For instance, unsupervised and adversarial pre-training methods have been employed to enhance the logical reasoning capabilities of language models \cite{pi2022logigan}. Using pre-training to inject numerical \cite{pi2022reasoning} and commonsense reasoning \cite{zhong2019improving} has also been recently explored. Huebner et.al have constructed pre-training paradigms using curriculum learning \cite{huebner-etal-2021-babyberta} to show the advantages of incremental learning. 

\subsection{Pre-training with limited data} 
Previous experiments show that pre-training data size is positively correlated with the syntactic capabilities of RoBERTa in terms of generalization and robustness \cite{perez2021much}. However, it has been discovered that model performance gains bring a high financial and environmental cost \cite{tay2021scale}. This justifies the appeal of small-scale pretraining with data limitations. There have also been explorations of how human-like data scales could improve our understanding of language acquisition and solidify current cognitive models \cite{Dupoux_2018}.

\begin{table}[!ht]
\resizebox{0.5\textwidth}{!}{%
    \centering
    \begin{tabular}{lccc}
    \hline
       Track &  Model & Competition Scores &  Perplexity\\ 
       ~ &  ~ & (Dynabench) & \\ 
        \hline
        Strict Small & Distilbert Epoch 20 & 0.62  & 86.283 \\ 
        ~ & Distilbert Epoch 60 & \textbf{0.65} & 17.278 \\ 
        ~ & RoBERTa Epoch 20 & 0.58 & 49.586 \\ 
        ~ & GPT2 Epoch 20 & 0.64 & 79.318 \\ 
        \hline
        ~ & Competition Max & 0.73 \\ 
        \hline
        Strict & Distilbert Epoch 20 & 0.66 & 39.427 \\ 
        ~ & Distilbert Epoch 60 & \textbf{0.71} & 10.332\\ 
        ~ & RoBERTa Epoch 20 & 0.63 & 27.566\\
        ~ & GPT2 Epoch 20 & 0.67 & 34.950\\ 
        \hline
        ~ & Competition Max & 0.81 \\ \hline
    \end{tabular}
    }
    \caption{Model scores on dynabench}
    \label{tab:cum_res}
\end{table}

\begin{table*}[!ht]
\resizebox{1\textwidth}{!}{%
   \centering
    \begin{tabular}{l|l|ccccccccccc}
    \hline
        Tasks & & \multicolumn{11}{c}{\begin{tabular}[c]{@{}c@{}}  Super GLUE \end{tabular}}  \\ \hline
        ~ & Model & CoLA & SST-2 & MRPC (F1) & QQP (F1) & MNLI & MNLI-mm & QNLI & RTE & BoolQ & MultiRC & WSC \\ \hline
        Strict Small & Majority label & 69.50 & 50.20 & \textbf{ 82.00} & 53.10 & 35.70 & 35.70 & 35.40 & 53.10 & 50.50 & 59.90 & 53.20 \\ 
        ~ & OPT-125m & 64.60 & 81.90 & 72.50 & 60.40 & 57.60 & 60.00 & 61.50 & 60.00 & 63.30 & 55.20 & 60.20 \\ 
        ~ & RoBERTa-base & \textbf{70.80} & \textbf{87.00} & 79.20 & 73.70 & 73.20 & \textbf{74.00} & \textbf{77.00} & \textbf{61.60} & 66.30 & \textbf{61.40} & 61.40 \\ 
        ~ & T5-base & 61.20 & 78.10 & 80.50 & 66.20 & 48.00 & 50.30 & 62.00 & 49.40 & 66.00 & 47.10 & 61.40 \\ 
        ~ & Distilbert Epoch 20 & 69.38 & 83.46 & 79.69 & 80.21 & 69.80 & 71.56 & 60.15 & 54.55 & 65.42 & 53.67 & 51.81 \\ 
        ~ & Distilbert Epoch 60 & 69.68 & 85.63 & 78.81 & \textbf{82.28} & \textbf{71.62} & 73.11 & 76.73 & 60.61 & \textbf{67.77} & 56.74 & 61.45 \\ 
        ~ & RoBERTa Epoch 20 & 65.55 & 81.30 & 79.71 & 76.37 & 65.16 & 65.82 & 62.73 & 56.57 & 62.38 & 44.91 & \textbf{61.45} \\ 
        ~ & GPT2 Epoch 20 & 69.58 & 83.07 & 75.47 & 73.13 & 63.88 & 65.95 & 59.84 & 56.57 & 64.45 & 58.38 & 46.99 \\ \hline
        Strict & OPT-125m & 73.70 & 86.60 & 82.10 & 77.80 & 70.10 & 71.90 & 80.10 & \textbf{67.70} & 66.00 & 61.10 & 59.00 \\ 
        ~ & RoBERTa-base & 75.90 & \textbf{88.60} & 80.50 & 78.50 & 68.70 & \textbf{78.00} & 82.30 & 51.50 & 59.90 & 61.30 & 61.40 \\ 
        ~ & T5-base & \textbf{76.30} & 88.00 & \textbf{85.90} & 79.70 & 71.50 & 74.00 & \textbf{83.10} & 60.60 & 69.00 & 62.40 & 60.20 \\ 
        ~ & Distilbert Epoch 20 & 69.48 & 86.22 & 62.98 & 83.81 & 73.44 & 74.97 & 79.00 & 60.61 & 67.91 & 62.98 & 44.58 \\ 
        ~ & Distilbert Epoch 60 & 74.78 & 87.01 & 81.40 & \textbf{84.37} & 74.95 & 75.27 & 80.97 & 55.56 & 65.56 & \textbf{65.83} & \textbf{61.45} \\ 
        ~ & RoBERTa Epoch 20 & 67.81 & 84.06 & 82.00 & 82.12 & 72.22 & 73.19 & 77.17 & 53.54 & 60.30 & 51.48 & 38.55 \\ 
        ~ & GPT2 Epoch 20 & 69.58 & 87.20 & 79.29 & 82.23 & \textbf{74.00} & 74.98 & 81.01 & 52.53 & \textbf{69.58} & 57.83 & 48.19 \\ \hline
    \end{tabular}
    }
    \caption{Results for the Super GLUE tasks}
    \label{tab:sglue}
\end{table*}
\begin{table*}[!ht]
\resizebox{\textwidth}{!}{%
   \centering
    \begin{tabular}{l|l|cccccccccccc}
    \hline
        Tasks & & \multicolumn{12}{c}{\begin{tabular}[c]{@{}c@{}}  Blimp \end{tabular}}  \\ \hline
        ~ & Model & Anaphor & Agr. & Binding & Control/ & D-N  & Ellipsis & Filler-Gap & Irregular  & Island  & NPI  & Quantifiers & S-V   \\
        ~ & ~ &  Agr. & Structure & Binding & Raising &Agr. &  & & Forms &  Effects &  Licensing &  & Agr.  \\
        \hline
       
        Strict Small & OPT-125m & 63.8 & \textbf{70.6} & 67.1 & 66.5 & 78.5 & 62 & 63.8 & 67.5 & 48.6 & 46.7 & 59.6 & 56.9 \\ 
        ~ & RoBERTa-base & 81.5 & 67.1 & 67.3 & \textbf{67.9} & \textbf{90.8} & 76.4 & 63.5 & 87.4 & 39.9 & \textbf{55.9} & \textbf{70.5} & 65.4 \\ 
        ~ & T5-base & 68.9 & 63.8 & 60.4 & 60.9 & 72.2 & 34.4 & 48.2 & 77.6 & 45.6 & 47.8 & 61.2 & 65 \\ 
        ~ & Distilbert Epoch 20 & 83.49 & 64.12 & 63.98 & 62.22 & 77.72 & 62.76 & 62.36 & 85.24 & 42.94 & 41.38 & 67.47 & 55.81 \\ 
        ~ & Distilbert Epoch 60 & \textbf{89.62} & 68.44 & 64.08 & 65.20 & 89.70 & \textbf{81.64} & 63.57 & \textbf{89.92} & 39.69 & 44.58 & 66.20 & \textbf{78.09} \\ 
        ~ & RoBERTa Epoch 20 & 84.76 & 60.54 & \textbf{67.97} & 60.69 & 56.47 & 52.25 & \textbf{65.48} & 64.53 & \textbf{54.22} & 52.51 & 52.42 & 66.63 \\ 
        ~ & GPT2 Epoch 20 & 81.24 & 72.56 & 67.81 & 67.43 & 86.98 & 59.82 & 67.72 & 84.38 & 52.62 & 51.76 & 58.14 & 64.12 \\ \hline
        Strict & OPT-125m & \textbf{94.9} & 73.8 & \textbf{73.8} & \textbf{72.2} & 93.1 & 80.5 & \textbf{73.6} & 80.8 & \textbf{57.8} & 51.6 & \textbf{74.5} & \textbf{77.3} \\ 
        ~ & RoBERTa-base & 89.5 & 71.3 & 71 & 67.1 & 93.1 & 83.8 & 68 & 89.6 & 54.5 & \textbf{66.3} & 70.3 & 76.2 \\ 
        ~ & T5-base & 66.7 & 61.2 & 59.4 & 59.8 & 53.8 & 49.1 & 70 & 75.5 & 43.6 & 45.6 & 34.2 & 53.2 \\ 
        ~ & Distilbert Epoch 20 & 92.43 & 67.06 & 67.66 & 65.27 & 94.38 & 87.24 & 65.42 & 85.04 & 42.86 & 50.43 & 67.41 & 66.25 \\ 
        ~ & Distilbert Epoch 60 & 94.68 & 70.39 & 68.39 & 68.25 & \textbf{96.39} & \textbf{89.03} & 68.69 & \textbf{90.08} & 45.59 & 64.67 & 70.20 & 72.32 \\ 
        ~ & RoBERTa Epoch 20 & 85.94 & 67.68 & 65.27 & 63.74 & 91.04 & 75.52 & 62.98 & 87.23 & 46.41 & 44.47 & 61.46 & 60.51 \\ 
        ~ & GPT2 Epoch 20 & 91.56 &  \textbf{74.88} & 73.21 & 69.22 & 91.89 &  75.52 & 71.91 &  75.32 & 55.04 & 51.20 &  66.13 & 67.19\\ \hline
    \end{tabular}
    }
    \caption{Results for the Blimp tasks}
    \label{tab:blimp}
\end{table*}
\begin{table*}[!ht]
\resizebox{\textwidth}{!}{%
   \centering
    \begin{tabular}{l|l|ccccc}
    \hline
        Tasks & & \multicolumn{5}{c}{\begin{tabular}[c]{@{}c@{}} Blimp Supplement Tasks \end{tabular}}  \\ \hline
        ~ & Model & Hypernym & QA Congruence (easy) & QA Congruence (tricky) & Subj.-Aux. Inversion & Turn Taking \\ \hline
        Strict Small & OPT-125m & 50.00 & 54.7 & 31.5 & 80.3 & 57.1 \\ 
        ~ & RoBERTa-base & 49.4 & 31.3 & 32.1 & 71.7 & 53.2 \\ 
        ~ & T5-base & 48 & 40.6 & 21.2 & 64.9 & 45 \\ 
        ~ & Distilbert Epoch 20 & 50.00 & 65.63 & \textbf{42.42} & \textbf{77.31} & 61.79 \\ 
        ~ & Distilbert Epoch 60 & 48.95 & \textbf{70.31} & 41.21 & 60.87 & 62.86 \\ 
        ~ & RoBERTa Epoch 20 & \textbf{51.28} & 48.44 & 31.52 & 53.86 & \textbf{66.07} \\ 
        ~ & GPT2 Epoch 20 & 47.44 & 48.44 & 45.45 & 72.41 & 62.86 \\ \hline
        Strict & OPT-125m & 46.3 & \textbf{76.50} & \textbf{47.9} & \textbf{85.3} & \textbf{82.9} \\ 
        ~ & RoBERTa-base & 50.8 & 34.4 & 34.5 & 45.6 & 46.8 \\ 
        ~ & T5-base & 51.1 & 45.3 & 25.5 & 69.2 & 48.9 \\ 
        ~ & Distilbert Epoch 20 & 48.26 & 64.06 & 40.61 & 81.53 & 65.36 \\ 
        ~ & Distilbert Epoch 60 & 48.95 & 73.44 & 47.88 & 83.43 & 65.36 \\ 
        ~ & RoBERTa Epoch 20 & \textbf{51.16} & 46.88 & 37.58 & 76.85 & 64.29 \\ 
        ~ & GPT2 Epoch 20 & 49.53 & 57.81 &  45.45 &   81.85 &  65.00\\ \hline
    \end{tabular}
    }
    \caption{Results for the Blimp supplemental tasks}
    \label{tab:blimp_sup}
\end{table*}

\begin{table*}[!ht]
\resizebox{\textwidth}{!}{%
   \centering
    \begin{tabular}{l|l|ccccccccccc}
    \hline
        Tasks & & \multicolumn{11}{c}{\begin{tabular}[c]{@{}c@{}} MSGS Tasks \end{tabular}}  \\ \hline
        
       ~ & Model & CR  & LC  & MV  & RP & SC  & CR\_LC & CR\_RTP & MV\_LC & MV\_RTP & SC\_LC & SC\_RP \\

       ~ & ~ & (Control) &  (Control) &  (Control) &  (Control) &  (Control) & &  &  &  &  &  \\ 
       \hline
        Strict-Small & OPT-125m & 86.40 & 86.10 & \textbf{99.80} & \textbf{100.00} & 94.30 & 66.50 & 67.00 & 66.50 & \textbf{67.60} & 80.20 & 67.50 \\ 
        ~ & RoBERTa-base & 84.10 & \textbf{100.00} & 99.40 & 93.50 & 96.40 & \textbf{67.70} & 68.60 & \textbf{66.70} & 68.60 & \textbf{84.20} & 65.70 \\ 
        ~ & T5-base & 78.40 & \textbf{100.00} & 72.70 & 95.50 & 94.40 & 66.70 & \textbf{69.70} & 66.60 & 66.90 & 73.60 & 67.80 \\ 
        ~ & Distilbert Epoch 20 & 79.22 & \textbf{100.00} & 97.17 & 98.57 & 96.36 & 66.53 & 66.71 & 66.61 & 67.47 & 67.89 & 67.58 \\ 
        ~ & Distilbert Epoch 60 & 81.68 & \textbf{100.00} & 98.61 & 99.14 & 95.66 & 67.24 & 66.72 & 66.61 & 67.03 & 67.76 & \textbf{68.27} \\ 
        ~ & RoBERTa Epoch 20 & 73.02 & \textbf{100.00} & 73.91 & 99.59 & 86.47 & 66.70 & 67.19 & 66.61 & 66.84 & 67.44 & 71.93 \\ 
        ~ & GPT2 Epoch 20 & \textbf{89.78} & 96.30 & 99.23 & \textbf{100.00} & \textbf{97.13} & 66.46 & 66.72 & 66.58 & 66.83 & 78.78 & 64.87 \\ \hline
        Strict & OPT-125m & \textbf{97.20} & 82.60 & \textbf{100.00} & 99.80 & 88.10 & 75.30 & 67.10 & 66.30 & 66.80 & 84.80 & 62.00 \\ 
        ~ & RoBERTa-base & 93.00 & \textbf{100.00} & \textbf{100.00} & \textbf{100.00} & 89.00 & 68.30 & 66.80 & 66.60 & \textbf{80.20} & 67.40 & 67.40 \\ 
        ~ & T5-base & 95.10 & \textbf{100.00} & \textbf{100.00} & 99.80 & 88.70 & \textbf{76.70} & \textbf{69.40} & \textbf{67.00} & 67.70 & 72.70 & 68.00 \\ 
        ~ & Distilbert Epoch 20 & 81.44 & \textbf{100.00} & 97.36 & 97.35 & 94.77 & 67.26 & 66.72 & 66.61 & 66.97 & 67.67 & 68.63 \\ 
        ~ & Distilbert Epoch 60 & 93.23 & \textbf{100.00} & 99.33 & 99.17 & 95.64 & 68.91 & 66.77 & 66.61 & 67.45 & 67.89 & 66.59 \\ 
        ~ & RoBERTa Epoch 20 & 84.63 & 97.38 & 92.12 & 98.15 & 95.54 & 66.47 & 66.59 & 66.41 & 66.05 & 68.17 & \textbf{72.78} \\ 
        ~ & GPT2 Epoch 20 & 95.35 & 76.53 & 99.55 & 99.83 & \textbf{96.76} & 67.21 & 68.46 & 66.78 & 66.70 & \textbf{91.90} & 65.90 \\ \hline
    \end{tabular}
    }
    \caption{Results for the MSGS tasks}
    \label{tab:msgs}
\end{table*}

\begin{table*}[!ht]
\resizebox{\textwidth}{!}{%
   \centering
    \begin{tabular}{l|l|cccc}
    \hline
        Tasks & & \multicolumn{4}{c}{\begin{tabular}[c]{@{}c@{}}  Age of Acquisition tasks (mean absolute deviation) \end{tabular}}  \\ \hline
        ~ & Model & Overall (591 words) & Nouns (322) & Predicates (167) & Function words (102) \\ \hline
        Strict Small  & OPT-125m & 2.03 & 1.98 & 1.81 & 2.57 \\ 
        ~ & RoBERTa-base & 2.06 & 1.99 & 1.85 & 2.65 \\ 
        ~ & T5-base & 2.04 & 1.97 & 1.82 & 2.64 \\ 
        ~ & Distilbert Epoch 20 & 2.06 & 2.00 & 1.84 & 2.65 \\ 
        ~ & Distilbert Epoch 60 & 2.09 & 2.00 & 1.84 & 2.76 \\ 
        ~ & RoBERTa Epoch 20 & 2.06 & 2.00 & 1.84 & 2.63 \\ 
        ~ & GPT2 Epoch 20 & 2.06 & 2.00 & 1.85 & 2.64 \\ \hline
        Strict & OPT-125m & 2.04 & 1.97 & 1.83 & 2.61 \\ 
        ~ & RoBERTa-base & 2.06 & 1.99 & 1.82 & 2.66 \\ 
        ~ & T5-base & 2.06 & 2.00 & 1.83 & 2.65 \\ 
        ~ & Distilbert Epoch 20 & 2.06 & 2.00 & 1.83 & 2.65 \\ 
        ~ & Distilbert Epoch 60 & 2.08 & 2.00 & 1.81 & 2.79 \\ 
        ~ & RoBERTa Epoch 20 & 2.06 & 2.00 & 1.84 & 2.62 \\ 
        ~ & GPT2 Epoch 20 & 2.04 & 1.98 & 1.81&2.60\\ \hline
    \end{tabular}
    }
    \caption{Results for the Age of Acquisition tasks}
    \label{tab:aoa}
\end{table*}
\section{Methodology}
\subsection{Models}
We use the simple-transformers library \cite{rajapakse2019simpletransformers} to pre-train the models below from scratch. The library uses the Huggingface trainer for pre-training. Note: We build new vocabularies for all models and limit the number of training epochs due to computational constraints in certain models.

\begin{itemize}\setlength\itemsep{0em}
    \item RoBERTa: We train the RoBERTa-base model \cite{RoBERTa} for comparison to the baseline given by the task organizers. This model is trained for 20 epochs on both datasets (strict and strict-small). The size of this model is roughly 125M parameters.
    \item DistilBert (uncased): Because this model \cite{sanh2020distilbert} is smaller (roughly 66M parameters) and quicker to pre-train, we additionally train it for 60 epochs. This allows us to explore the impact of more training epochs on performance. 
    \item GPT2: We include a decoder-based architecture \cite{gpt2} in our pre-training to explore the impact of architecture type on the evaluation tasks. This model has a similar size to RoBERTa (117M parameters). We train it for 20 epochs due to computational constraints.
\end{itemize}
 All of the checkpoints for the three architectures and the two tracks are uploaded on Huggingface \cite{huggingface}. \textbf{Hyperparameters:} We perform a grid search over the hyperparameters for all three architecture types. We use a  subset of 0.5 GB of the training data for the search. The learning rate ranges from 5e-5 to 4e-4 across the searches, with weight decay in place but no early stopping mechanisms.
 
\section{Results}

Table \ref{tab:cum_res} shows the results obtained from the dynabench submission portal. The individual results for each of the tasks in different benchmarks are available in Tables \ref{tab:sglue}, \ref{tab:blimp}, \ref{tab:blimp_sup}, \ref{tab:msgs}, \ref{tab:aoa}. Looking at these tables, we observe the following patterns:
\begin{enumerate}
    \item We see that training for more epochs leads to better overall performance (compare 20 and 60 epochs of DistilBert in Table \ref{tab:cum_res}).
    \item Variation among architecture types exists when limiting the training to the same number of epochs, but it is difficult to identify a definitively better architecture.
    \item Tables \ref{tab:sglue}, \ref{tab:blimp}, \ref{tab:blimp_sup}, \ref{tab:msgs}, and \ref{tab:aoa} show that pre-training (RoBERTa) is not robust to initialization, and the competition scores would greatly benefit from a warm-up or a grid search over different hyper-parameters.
    \item In most cases, the pre-training improves performance over the majority label in the Super GLUE tasks.
    \item Tables \ref{tab:rob_epochs}, \ref{tab:dis_epochs} shows that the performance on the BLIMP tasks becomes better with more training epochs. While this is orthogonal to wisdom performance saturates at one epoch\cite{biderman2023pythia}. Our results hint that training saturation or stability may be a function of model size divided by the number of tokens seen.
\end{enumerate}
\begin{table*}[!ht]
\resizebox{\textwidth}{!}{%
   
    \centering
    \begin{tabular}{l|cccccccccccccccccccc}
    \hline
     & \multicolumn{20}{c}{\begin{tabular}[c]{@{}c@{}}  Epochs \end{tabular}}  \\ \hline
        BLIMP Tasks & 1 & 2 & 3 & 4 & 5 & 6 & 7 & 8 & 9 & 10 & 11 & 12 & 13 & 14 & 15 & 16 & 17 & 18 & 19 & 20 \\
      
        \hline
        Anaphor Agr. & 50.77 & 70.81 & 82.21 & 79.75 & 84.46 & 84.00 & 86.71 & 88.14 & 87.27 & 87.88 & 87.42 & 87.32 & 83.54 & 86.91 & 86.30 & 84.76 & 85.22 & 85.99 & 85.22 & 85.94 \\ 
        Agr. Str. & 58.69 & 59.81 & 60.46 & 60.83 & 59.80 & 60.39 & 60.43 & 60.84 & 61.02 & 60.60 & 59.69 & 62.08 & 63.70 & 65.22 & 65.39 & 66.56 & 67.54 & 67.34 & 67.92 & 67.68 \\ 
        Binding & 64.43 & 68.06 & 64.84 & 65.32 & 64.31 & 67.17 & 64.31 & 66.00 & 63.92 & 64.10 & 64.00 & 59.45 & 61.52 & 62.60 & 62.72 & 63.67 & 64.54 & 65.21 & 65.18 & 65.27 \\
        Control Rais. & 59.17 & 59.19 & 58.17 & 58.24 & 59.06 & 59.01 & 59.79 & 59.70 & 59.70 & 60.01 & 58.51 & 62.22 & 60.25 & 59.41 & 60.87 & 62.88 & 62.79 & 63.57 & 63.85 & 63.74 \\ 
        Det-N Agr. & 51.47 & 56.72 & 58.87 & 59.57 & 58.63 & 58.62 & 58.18 & 60.09 & 60.16 & 59.06 & 59.63 & 63.47 & 71.69 & 80.95 & 85.20 & 87.18 & 88.49 & 90.51 & 90.98 & 91.04 \\ 
        Ellipsis & 37.93 & 44.05 & 46.54 & 50.87 & 50.98 & 54.27 & 56.12 & 58.43 & 59.70 & 61.61 & 55.02 & 51.96 & 54.62 & 65.47 & 68.30 & 73.85 & 72.92 & 75.64 & 75.23 & 75.52 \\ 
        Filler Gap & 69.39 & 64.53 & 66.48 & 63.66 & 65.38 & 61.95 & 61.31 & 61.95 & 61.84 & 65.39 & 60.58 & 61.86 & 60.21 & 62.14 & 61.47 & 61.80 & 62.40 & 63.35 & 63.38 & 62.98 \\ 
        Hypernym & 53.84 & 49.77 & 52.09 & 51.74 & 48.02 & 48.49 & 48.72 & 50.12 & 49.88 & 51.28 & 50.35 & 50.70 & 48.84 & 50.35 & 51.28 & 49.77 & 51.51 & 50.47 & 51.86 & 51.16 \\ 
        Irr. Forms& 45.90 & 59.75 & 60.87 & 61.32 & 63.66 & 59.54 & 65.24 & 63.36 & 64.83 & 64.99 & 69.97 & 78.07 & 76.39 & 81.53 & 86.92 & 86.87 & 88.96 & 88.85 & 87.74 & 87.23 \\ 
        Island Effects & 53.66 & 42.68 & 56.20 & 55.53 & 54.67 & 44.66 & 48.28 & 51.91 & 49.93 & 52.88 & 47.31 & 45.22 & 51.76 & 50.67 & 47.83 & 48.92 & 46.82 & 46.38 & 45.40 & 46.41 \\ 
        NPI Lic.& 34.89 & 43.58 & 35.71 & 46.96 & 40.81 & 43.41 & 43.29 & 42.97 & 39.60 & 44.02 & 37.25 & 38.23 & 39.20 & 40.80 & 40.10 & 43.77 & 44.05 & 44.46 & 43.65 & 44.47 \\ 
        QA\_cong. Easy & 34.38 & 35.94 & 39.06 & 42.19 & 39.06 & 35.94 & 40.63 & 40.63 & 37.50 & 37.50 & 37.50 & 46.88 & 56.25 & 57.81 & 56.25 & 50.00 & 45.31 & 45.31 & 48.44 & 46.88 \\ 
        QA\_cong. Tricky& 38.18 & 34.55 & 32.73 & 31.52 & 31.52 & 30.91 & 30.30 & 29.70 & 28.48 & 28.48 & 27.27 & 24.85 & 23.64 & 30.30 & 33.94 & 32.73 & 35.76 & 35.15 & 38.18 & 37.58 \\ 
        Quantifiers & 37.92 & 38.15 & 36.22 & 40.96 & 43.07 & 41.96 & 49.00 & 50.05 & 43.07 & 51.91 & 51.52 & 66.95 & 65.35 & 67.85 & 60.72 & 64.30 & 61.98 & 64.09 & 61.39 & 61.46 \\ 
        S-Aux Inv. & 74.68 & 75.04 & 69.85 & 68.87 & 63.19 & 52.74 & 55.40 & 50.65 & 47.89 & 48.65 & 52.28 & 62.94 & 66.85 & 72.82 & 70.36 & 77.92 & 73.82 & 76.38 & 75.43 & 76.85 \\ 
        S-V Agr.& 49.67 & 50.89 & 52.10 & 51.91 & 52.34 & 53.68 & 53.80 & 54.47 & 55.54 & 55.83 & 54.87 & 52.72 & 54.27 & 57.58 & 58.10 & 58.70 & 60.09 & 60.31 & 60.23 & 60.51 \\ 
        Turn-Taking & 59.64 & 59.64 & 60.00 & 59.29 & 61.43 & 60.36 & 58.57 & 59.64 & 58.93 & 58.57 & 61.07 & 63.57 & 64.64 & 67.14 & 63.57 & 63.57 & 64.29 & 63.93 & 64.64 & 64.29 \\
        \hline
    \end{tabular}
    }
    \caption{Results for the BLIMP tasks across different epochs of the RoBERTa-base model architecture for the strict (100M token) track. }
    \label{tab:rob_epochs}
\end{table*}

\begin{table*}[!ht]
\resizebox{\textwidth}{!}{%
   
    \centering
    \begin{tabular}{l|ccccccccccccc}
    \hline
     & \multicolumn{13}{c}{\begin{tabular}[c]{@{}c@{}}  Epochs \end{tabular}}  \\ \hline
        Behavior/ Model +Epoch & 1 & 5 & 10 & 15 & 20 & 25 & 30 & 35 & 40 & 45 & 50 & 55 & 60  \\
      
        \hline
        Anaphor Agr. & 46.57 & 82.87 & 89.88 & 91.21 & 92.43 & 93.10 & 94.07 & 94.17 & 95.19 & 94.94 & 94.58 & 94.43 & 94.68 \\ 
        Agr. Str. &  58.06 & 59.71 & 61.78 & 65.69 & 67.06 & 68.02 & 70.05 & 69.07 & 69.67 & 70.55 & 70.49 & 70.27 & 70.39 \\ 
        Binding & 59.65 & 65.24 & 63.15 & 67.14 & 67.66 & 66.93 & 68.48 & 66.55 & 69.07 & 68.76 & 68.95 & 68.27 & 68.39 \\
        Control Rais. & 58.33 & 58.93 & 60.01 & 64.14 & 65.27 & 66.00 & 65.91 & 67.12 & 67.30 & 67.41 & 68.10 & 67.87 & 68.25 \\ 
        Det-N Agr. & 50.76 & 60.30 & 70.41 & 92.16 & 94.38 & 95.24 & 95.94 & 95.97 & 96.34 & 96.14 & 96.27 & 96.37 & 96.39 \\ 
        Ellipsis & 37.53 & 54.16 & 55.08 & 81.99 & 87.24 & 86.49 & 86.20 & 89.32 & 89.32 & 89.38 & 88.57 & 88.86 & 89.03 \\ 
        Filler Gap & 70.23 & 64.89 & 58.56 & 62.06 & 65.42 & 64.74 & 66.64 & 67.49 & 67.54 & 67.24 & 69.00 & 68.88 & 68.69 \\ 
        Hypernym & 51.40 & 50.23 & 50.70 & 48.84 & 48.26 & 50.00 & 48.60 & 51.40 & 50.23 & 50.00 & 49.77 & 48.49 & 48.95 \\ 
        Irr. Forms& 56.39 & 65.24 & 87.38 & 85.55 & 85.04 & 86.92 & 88.50 & 89.16 & 88.85 & 88.85 & 89.72 & 89.72 & 90.08 \\ 
        Island Effects & 46.52 & 44.62 & 48.09 & 45.52 & 42.86 & 45.07 & 43.20 & 46.49 & 44.81 & 43.80 & 44.39 & 45.44 & 45.59 \\ 
        NPI Lic.& 53.23 & 46.90 & 41.06 & 46.67 & 50.43 & 55.25 & 58.56 & 57.39 & 61.69 & 64.36 & 64.09 & 64.15 & 64.67 \\ 
        QA\_cong. Easy & 31.25 & 43.75 & 59.38 & 67.19 & 64.06 & 68.75 & 70.31 & 73.44 & 75.00 & 70.31 & 70.31 & 73.44 & 73.44 \\ 
        QA\_cong. Tricky& 333.33 & 22.42 & 23.03 & 35.15 & 40.61 & 42.42 & 46.06 & 43.03 & 44.24 & 41.82 & 46.06 & 46.06 & 47.88 \\ 
        Quantifiers & 54.87 & 69.55 & 62.31 & 65.43 & 67.41 & 70.40 & 70.50 & 72.82 & 70.74 & 70.63 & 70.25 & 70.81 & 70.20 \\ 
        S-Aux Inv. & 58.45 & 65.77 & 73.65 & 79.21 & 81.53 & 81.19 & 81.85 & 81.75 & 83.17 & 82.80 & 83.63 & 82.53 & 83.43 \\ 
        S-V Agr.& 48.93 & 54.60 & 55.56 & 62.06 & 66.25 & 68.24 & 70.82 & 70.05 & 71.64 & 72.50 & 71.73 & 72.41 & 72.32 \\ 
        Turn-Taking & 59.29 & 60.71 & 65.36 & 64.29 & 65.36 & 63.93 & 64.64 & 65.36 & 65.36 & 65.00 & 66.07 & 65.00 & 65.36 \\
        \hline
    \end{tabular}
    }
    \caption{Results for the BLIMP tasks across different epochs of the DistilBERT-base model architecture for the strict (100M token) track. }
    \label{tab:dis_epochs}
\end{table*}

\section{Conclusions}

We pre-train popular LLM architectures on the kind of textual data seen by children when they are around 13 years old. We show that pre-training paradigms like Masked Language Modeling or Causal Language Modeling lead to only minor variations. Our results show that models are not robust to the initialization of weights. Our work provides each and every checkpoint of the model architectures on Huggingface to facilitate future research. All checkpoints can be found \textbf{\href{https://huggingface.co/Raj-Sanjay-Shah}{here.}}

\section{Limitations}
Our work trains some of the popular Language Model architectures on human-like scaled-down training data, it does not introduce new training methodologies or architectures which may be better suited for such tasks. Furthermore, our work does not exhaustively cover different model types in the literature. Our results are preliminary as they do not account for all possible confounds.

\section{Ethical Considerations}
All researchers in this study have active responsible code of conduct in research certifications. The models shared on Huggingface have the same risks associated with any other Large Language Model. Researchers in this study have tried to be mindful of the environment while doing the pre-training runs and hope that publically available checkpoints will help other researchers avoid computation and environmental costs associated with repeat pre-training.
\section{Computational Resources}
The models are trained on Nvidia-RTX 2080 GPUs with 12 GB RAM. The models are trained for nearly 975 GPU hours.
\bibliography{emnlp2023}
\bibliographystyle{acl_natbib}

\appendix

\end{document}